\documentclass[12pt,twoside]{article}
\usepackage{times}
\usepackage[utf8]{inputenc}
\usepackage[T1]{fontenc}
\usepackage{graphicx}

\usepackage{booktabs}
\usepackage{multirow}
\usepackage{tabularx}
\usepackage{tabu}
\usepackage[english]{babel}
\usepackage{lipsum}
\usepackage{hyperref}
\usepackage{tabularray}

\usepackage{enumitem}
\setlist{nolistsep}

\usepackage{caption} 
\usepackage{wrapfig}

\usepackage{xspace}
\newcommand{\camemberta}{CamemBERTa\xspace}

\usepackage{siunitx}

\usepackage{taln2023}

\title{Towards a Robust Detection of Language Model Generated Text:\\ Is ChatGPT that Easy to Detect?}

\author{Wissam Antoun\quad Virginie Mouilleron \quad Benoît Sagot\quad Djamé Seddah\\
  {\small
    Inria, Paris, France \\ 
    \texttt{
     firstname.lastname@inria.fr \\ 
}}}

\begin{document}

\maketitle

\abstract{
  Recent advances in natural language processing (NLP) have led to the development of large language models (LLMs) such as ChatGPT.
This paper proposes a methodology for developing and evaluating ChatGPT detectors for French text, with a focus on investigating their robustness on out-of-domain data and against common attack schemes. 
The proposed method involves translating an English dataset into French and training a classifier on the translated data. 
Results show that the detectors can effectively detect ChatGPT-generated text, with a degree of robustness against basic attack techniques in in-domain settings. 
However, vulnerabilities are evident in out-of-domain contexts, highlighting the challenge of detecting adversarial text. 
The study emphasizes caution when applying in-domain
testing results to a wider variety of content. We provide our translated datasets and models as open-source resources.\footnote{\href{https://gitlab.inria.fr/wantoun/robust-chatgpt-detection}{https://gitlab.inria.fr/wantoun/robust-chatgpt-detection}}

}

\resume{Vers une détection robuste de texte généré par un modèle de langue~: ChatGPT est-il si facile à détecter~?}{
Les récents progrès en traitement automatique des langues (TAL) ont conduit au développement de grands modèles de langage ({\em Large Language Models}, LLM) tels que ChatGPT.
Cet article propose une méthodologie pour développer et évaluer des modèles de détection de contenus en français produits par ChatGPT, en mettant l'accent sur leur robustesse face à des données hors domaine et face à des attaques classiques.
La méthode proposée consiste à traduire un ensemble de données anglaises en français et à entraîner un classificateur sur les données traduites.
Les résultats montrent que les détecteurs peuvent efficacement détecter le texte généré par ChatGPT, avec un bon niveau de robustesse contre les techniques usuelles d'attaque sans changement de domaine.
Cependant, les résultats sont moins bons dans les contextes hors domaine, soulignant le défi que constitue toujours de contenus adversariaux.
Notre étude souligne l'importance de rester prudent lorsque l'on cherche à généraliser des résultats obtenus sans changement de domaine à une plus grande variété de contenus. Tous nos jeux de données et nos modèles sont distribués librement.
}

\motsClefs
  {ChatGPT, génération de texte, détection de texte généré par machine, robustesse}
  {ChatGPT, text generation, detection of machine-generated text, robustness}

\section{Introduction}

Advances in natural language processing (NLP) have been driven mainly by scaling up the size of pre-trained language models, along with the amount of data and compute required for training~\cite{raffel2019exploring,radford2019language,rae2021scaling,fedus2021switch,hoffmann2022training}.
OpenAI recently released ChatGPT, a text generation model with conversational capabilities.
The model is based on GPT3.5 which is a version of GPT3~\citep{brown2020language} first fine-tuned on code then fine-tuned using Reinforcement Learning from Human Feedback (RLHF)~\citep{christiano2017deep,Stiennon2020learning}, a method previously demonstrated by OpenAI with InstructGPT~\citep{ouyang2022training}.
This fine-tuning process contributes not only to the model's knowledge but also simplifies the model's interface compared to GPT3,  which necessitated substantial \textit{prompt engineering} to achieve satisfactory outcomes, and hence facilitating the extraction and application of that built-in knowledge.

As a result of these significant performance improvements, ChatGPT and other large language models have gained much popularity in the media and in the social context, often without fully understanding the underlying limitations of the models -- e.g., the possibility of generating hateful, hateful, toxic, or disrespectful content~\citep{bender2021onthedangers,mcguffie2020radicalization,weidinger2021ethical}.
Another potential misuse of LLMs or ChatGPT is industrializing radicalization and harmful propaganda which poses a significant and unconventional threat to civil society.

In response to the mounting concerns surrounding potential misuse, numerous researchers are now exploring various strategies to mitigate associated risks. 
For example, some have proposed watermarking techniques to trace the origin of generated text\footnote{
At the time of writing, OpenAI was working on a tool to statistically watermark text generated by GPT-like models according to Scott Aaronson, a guest researcher at OpenAI~\href{https://scottaaronson.blog/?p=6823}{https://scottaaronson.blog/?p=6823}
}, while others are developing methods to detect and flag text generated by these models.
Of particular interest, a recent study by \citet{guo2023close} investigated the text generation capabilities of ChatGPT and its proximity to human-generated text. 
To create a dataset of ChatGPT-generated text, the authors leveraged pre-existing question-answering datasets in both English and Chinese, using the questions as prompts to generate responses from the model.
In addition, the authors conducted a linguistic analysis to compare the output generated by ChatGPT with human-written text, and they also developed a detector to distinguish between ChatGPT-generated text and human-written text by fine-tuning a separate language model on a dataset containing both types of text.

The aim of this research is to explore the development of ChatGPT detectors in multiple languages, along with evaluating their robustness on out-of-domain text, we selected French as the language of interest. 
Therefore, we propose a methodology that involves translating the English dataset into French and subsequently training a classifier on the translated data. 
We conducted a series of evaluations in a monolingual and multilingual setting on both in-domain and out-of-domain data.
The in-domain data consisted of text generated by ChatGPT using prompts related to the topics covered in the training dataset. 
The out-of-domain data included text generated in French by ChatGPT and Bing, a search engine powered by ChatGPT\footnote{\href{https://blogs.microsoft.com/blog/2023/02/07/reinventing-search-with-a-new-ai-powered-microsoft-bing-and-edge-your-copilot-for-the-web/}{https://blogs.microsoft.com/blog/2023/02/07/reinventing-search-with-a-new-ai-powered-microsoft-bing-and-edge-your-copilot-for-the-web/}}, which has access to a broader range of internet content and may generate text on a wider range of topics than ChatGPT.
Given that \citet{wolff2020attacking} demonstrated the vulnerability of BERT-based detectors for GPT-2 against basic attack schemes, such as substituting characters with homoglyphs or misspelled words, we also evaluated the robustness of our models against these types of attacks.
Furthermore, we hypothesize that the detector models we trained rely heavily on the didactic response style of ChatGPT to distinguish between human-generated content.

The contributions of this study can be summarized as follows:
\begin{itemize}
    \item We build upon the work of \citet{guo2023close} and propose a methodology to develop ChatGPT detectors in multiple languages, focusing on French as a case study.
    \item We evaluated the performance of the ChatGPT detectors in both a monolingual and multilingual setting. Specifically, we trained and evaluated our models on both the English and French datasets, as well as on a combined dataset containing both languages.
    \item We investigate the generalizability of our detector by testing its performance on out-of-domain data.
    \item We evaluate the robustness of our models against common attack schemes, such as substituting characters with homoglyphs or misspelled words. This is an important aspect to consider in the deployment of ChatGPT detectors, as attackers may attempt to evade detection by modifying the generated text in subtle ways.
    \item We investigated the reliance of detector models on ChatGPT's didactic response style for distinguishing between human-generated content.
    \item We release all translated datasets and models as open source to encourage further research in this area and to enable others to replicate our experiments.
\end{itemize}

Overall, our work contributes to the growing body of research on developing and evaluating ChatGPT detectors, with a focus on multilinguality, generalizability, and robustness to attacks. Our findings have practical implications for the use of ChatGPT detectors in various settings, including social media platforms, online forums, and chatbots, where the detection of harmful content is critical for maintaining a safe and respectful online environment.
\section{Related Works}
\label{sec:related-works}

\subsection{Large Language Models}
The race to scale up language models to new heights has been a hot topic in recent years. 
Researchers and tech companies have been competing to develop larger and more powerful models, often breaking records for model size and performance.
The trend began with OpenAI's GPT-2~\citep{radford2018improving}, which was released in 2019 and featured 1.5 billion parameters. 
This was quickly followed by Megatron~\citep{shoeybi2019megatron}, a 8.3 billion parameter model, displaying steadily increasing superior zero-shot language model performance, and T5~\citep{raffel2019exploring}, an 11 billion parameter encoder-decoder model which advanced transfer learning and performance on several closed-book question-answering tasks.
The release of GPT-3~\citep{brown2020language}, and PaLM~\citep{chowdhery2022palm} represented a major milestone in the race to scale up language models, with their unprecedented 175 and 540B billion parameters.
Scaling models to such massive scales ``unlocks'' new emergent capabilities as shown in ~\citet{chowdhery2022palm}.
In November 2022, OpenAI released ChatGPT, a conversational language model based on GPT-3.5 fine-tuned using Reinforcement Learning from Human Feedback (RLHF)~\citep{christiano2017deep,Stiennon2020learning}, a method previously demonstrated by OpenAI with InstructGPT~\citep{ouyang2022training}.

\subsection{Detecting Synthetic Text}
Detecting synthetically generated text is one of the defense mechanisms against harm caused by LLMs.
One of the first major explorations of this topic was conducted following the release of GROVER~\cite{zellers2019defending} a fake news generator and detector.
Since this approach has been shown to work quite well, and as part of their model release strategies~\citep{solaiman2019release}, OpenAI also released a GPT2 detector based on a fine-tuned RoBERTa model~\citep{liu2019roberta}, later \citet{fagni2021tweepfake} demonstrated the performance of another RoBERTa-based detector on machine-generated tweets, \citet{uchendu-etal-2020-authorship} also used RoBERTa to spot news generated by several language models, while \citet{antoun-etal-2021-aragpt2} created an ELECTRA-based model~\cite{antoun-etal-2021-araelectra} to spot articles generated by their AraGPT2. 
The authors stated that the success of their method was due to the model being pre-trained on the exact same dataset as AraGPT2, and also due to the replaced-token detection pre-training objective (RTD)~\cite{clark2020electra}, which bears a resemblance to the synthetic text detection objective.
\citet{nguyen2021machine} proposed a detector that uses the text similarity with round-trip translation (TSRT), to detect a machine-translated text from a never before seen translator, and achieved 86.9\% detection accuracy.
On the other hand, \citet{wolff2020attacking} showed that the RoBERTa GPT-2 detector is vulnerable against simple attack schemes such as substituting characters with homoglyphs or misspelled words.
In these cases, the detector's recall went down from 97\% to 0.26\% and 22.68\% respectively.
In order to further enhance the accuracy of detecting manipulated news articles that may deceive readers, \citet{jawahar-etal-2022-automatic} proposed a neural network-based detector that uses factual knowledge via graph convolutional neural network to distinguish between human-written news articles and manipulated news articles that mislead readers.

Following the release of ChatGPT, and with the recognition of the potential risks posed by this highly-capable model, there has been a surge of investigations into methods for detecting ChatGPT-generated text, which led to the commercial release of multiple detector products.
However, as the methods used in these products are often not publicly verifiable, this study focuses solely on the academic literature surrounding detection methods.
\citet{mitchell2023detectgpt} proposes a new method called DetectGPT, which leverages negative curvature regions of the model's log probability function and does not require training a separate classifier or watermarking generated text, resulting in a more discriminative approach than existing zero-shot methods for model sample detection.
Notably, \citet{guo2023close} created a dataset of ChatGPT responses to queries from diverse sources in English and Chinese.
The authors investigated the linguistic and stylistic differences between human and ChatGPT-generated text, in addition to training a variety of classifiers of which a finetuned pretrained language model turned out to be the best.

\section{Methodology}
\subsection{Data Collection}

To train and evaluate our ChatGPT detectors, we leveraged the Human ChatGPT Comparison Corpus (HC3) created by \citet{guo2023close} which contains both human-written and ChatGPT-generated text in English and Chinese.
We primarily focus on the English portion of the dataset as machine translation performs optimally on it.
The dataset consists of 24,322 human-written questions and 58546 answers sourced primarily from ELI5~\cite{reddit-eli5_lfqa}, WikiQA~\citep{yang2015wikiqa}, Crawled Wikipedia, Medical Dialog dataset~\citep{chen2020MedDialog-en-zh}, and FiQA~\citep{fiqa-2018}.
The authors generated ChatGPT responses using OpenAI's web application,\footnote{\href{https://chat.openai.com/chat}{https://chat.openai.com/chat}} automating the input of questions and scraping the answers with the help of automation testing tools, for a total of 26903 ChatGPT-generated answers.

To create a French dataset, we translated the English dataset using the Google Cloud Translation API.
We then split the dataset into three splits train, validation, and test, by first selecting 710 balanced question and answer pairs to be validated, manually annotated\footnote{The detailed annotation guideline will be publicly released with our dataset.} and to serve as our test set. We split the rest in an 80/20 split to get the training and validation set.

Furthermore, to assess the ChatGPT detectors' ability to generalize, we manually compiled out-of-domain test data by means of:
\begin{itemize}
    \item Manually collecting 113 ChatGPT French responses to high-quality translated questions from the test set, referred to as the \textbf{ChatGPT-Native} set.
    \item Using Bing, we manually collect 106 French responses to high-quality translated questions from the test set which we refer to as the \textbf{BingGPT}. Given that BinGPT includes source citations in its output, we remove these artifacts from the data (as well as all of its self-referring mentions).
    \item Randomly sampling 4454 French question-answer pairs from the French subset of the Multilingual FAQ Dataset (MFAQ)~\citep{de-bruyn-etal-2021-mfaq}, known as the \textbf{FAQ-Rand} set.
    \item Since the French FAQ data featured in the MFAQ dataset could be machine translated, we create a smaller set from the French FAQ data featured by filtering for .gouv domains, named the \textbf{FAQ-Gouv} set.
    \item 1235 sentences from The French Treebank test set, corpus from Le Monde~\cite{abeille-etal-2000-building} articles, which we denote as the \textbf{FTB} set.
    \item Moreover, in order to investigate our hypothesis that the detector relies heavily on the style of ChatGPT and Bing answers to distinguish between human-generated content, we created an additional set of responses to 61 questions. These responses were crafted as ``open-book'' answers with the same style as those provided by ChatGPT and Bing, resulting in a set of responses that we refer to as the \textbf{Adversarial} set.
\end{itemize}

\subsection{Detector Architecture}
Our approach fine-tunes pre-trained transformer-based models on our binary classification dataset.

For English, we used two pre-trained transformer models: RoBERTa~\citep{liu2019roberta} and ELECTRA~\citep{clark2020electra}. 
RoBERTa is a variant of BERT~\citep{devlin-etal-2019-bert}, trained using masked language modeling.
ELECTRA, on the other hand, introduced a new training objective, Replaced Token Detection (RTD), that replaces tokens in the input sequence with tokens generated by another model and then requires the discriminator to distinguish between the replaced and original tokens.
We hypothesize that this objective should improve performance since the RTD objective greatly resembles the machine-generated text detection objective.
 
For French, we used two pre-trained transformer models: CamemBERT~\citep{martin-etal-2020-camembert} and \camemberta~\citep{antoun2023data}.
CamemBERT is a RoBERTa model trained from scratch on French text, while \camemberta is based on the DeBERTaV3~\citep{he2021debertav3} architecture and trained from scratch on French text using RTD.

For the multilingual setting, we only fine-tune XLM-R~\citep{conneau-etal-2020-unsupervised}, a multilingual RoBERTa model with supports for 100+ languages.\footnote{We also tested mDeBERTa~\citep{he2021debertav3} but it wasn't converging in any of our hyper-parameter tuning experiments.}

\section{Experimental Methodology and Results}
\subsection{Experiment Design}
Motivated by \citet{guo2023close}, and given that the HC3 dataset comprises question/answer pairs, we investigated three distinct methods for generating dataset examples:
\begin{itemize}
    \item Jointly incorporating the question and answer into the model input, which we refer to as the \textbf{qa} subset.
    \item Using only the full answer text, which we refer to as the \textbf{full} subset.
    \item Splitting the answer text into sentences, resulting in shorter text segments and producing 455,320 training examples and 114,117 validation examples. We refer to this subset as the \textbf{sentence} subset.
\end{itemize}
To test the robustness of our approach against adversarial attacks, we add misspellings and simulate homoglyph substitution on the \textbf{full} subset of the test sets, using the \textit{nlaug}~\citep{ma2019nlpaug} library.

Regarding our choices of training hyperparameters, we maintain a fixed batch size of 32, adopt a linear scheduler with a warmup ratio of 0.1\%, and restrict our learning rate tuning to a range between $10^{-5}$ and $5.10^{-5}$ with a step size of $10^{-5}$.
Our model is trained for 5 epochs, and we report the results averaged over 5 distinct random seeds for all in-domains results. For the out-of-domains experiments, we used the best models.

\subsection{Results}
\subsubsection{In Domain}
Table~\ref{tab:allmodels} presents the results obtained from hyperparameter tuning. 
Notably, both evaluated French models demonstrated exceptional performance, and consistent stability evidenced by the low standard deviation scores.
However, the scores for French models were comparatively lower than the English models, indicating the impact of translation on model performance.
Our findings suggest that the performance of models trained on the QA and Full subset significantly deteriorates when assessed on short-length or sentence data, indicated also by the high standard deviation scores.
Conversely, models trained on sentences exhibit a relatively consistent performance across all subsets. 
Considering the overall highest performance on the \textbf{Full} subset, we opted to conduct subsequent experiments with the \camemberta and RoBERTa models trained on the Full subset.

Furthermore, Table~\ref{tab:detailedf1score} displays a detailed breakdown of the scores obtained from the \textbf{Full} subset. 
Notably, the models consistently achieve high recall scores in identifying ChatGPT across all tested languages. 
Additionally, the inclusion of misspellings and homoglyph substitutions improves the models' ability to detect human-written text while slightly reducing their performance for machine-generated text. 
The multilingual model XLM-R demonstrates superior and more resilient performance on both the French and English test sets, exhibiting increased robustness against adversarial attacks.
When compared to a native French-speaking human linguist, the trained model accurately identifies ChatGPT-generated content with higher accuracy, while the human linguist achieves a similar human detection score.

\begin{table}[ht]
\resizebox{\columnwidth}{!}{%
\begin{tabular}{@{}l|ccc|ccc|ccc@{}}
\toprule
\multicolumn{1}{r|}{Train} & \multicolumn{3}{c|}{QA} & \multicolumn{3}{c|}{Full} & \multicolumn{3}{c}{Sentence} \\
\multicolumn{1}{r|}{Test}  & QA  & Full  & Sentence & QA   & Full  & Sentence  & QA    & Full    & Sentence   \\ \midrule  \midrule 
\multicolumn{10}{c}{\em French}\\ \midrule 
CamemBERT  & \textbf{98.37{\small$\pm$0.5}} & 97.79{\small$\pm$0.4} & \textbf{40.20{\small$\pm$8.6}} &  \textbf{92.43{\small$\pm$1.2}} & \textbf{98.44{\small$\pm$0.4}} &  25.08{\small$\pm$4.7} & \textbf{93.48{\small$\pm$5.2}} & 96.41{\small$\pm$0.6} & 90.27{\small$\pm$0.3} \\
\camemberta & 98.23{\small$\pm$0.3} & \textbf{98.48{\small$\pm$0.3}} &  32.00{\small$\pm$6.3} &  90.13{\small$\pm$1.0} &  \textbf{98.49{\small$\pm$0.4}} &  \textbf{29.11{\small$\pm$3.6}} & 81.82{\small$\pm$3.4} & \textbf{96.71{\small$\pm$0.1}} &  \textbf{91.18{\small$\pm$0.2}} \\ \midrule \midrule
\multicolumn{10}{c}{\em English}\\ \midrule 
RoBERTa    &  \textbf{99.88{\small$\pm$0.03}} &  \textbf{98.91{\small$\pm$0.2}} & 51.23{\small$\pm$7.6} &  \textbf{98.58{\small$\pm$0.7}} &  \textbf{99.86{\small$\pm$0.03}} &  \textbf{66.93{\small$\pm$5.4}} &  71.10{\small$\pm$19.4} & \textbf{99.39{\small$\pm$0.07}} &  \textbf{98.17{\small$\pm$0.1}} \\
ELECTRA    &  99.27{\small$\pm$0.2} &  \textbf{99.07{\small$\pm$0.2}} & \textbf{65.23{\small$\pm$8.3}} &  96.24{\small$\pm$0.7} &  99.35{\small$\pm$0.1} &  43.82{\small$\pm$9.1} &  \textbf{93.57{\small$\pm$1.9}} &  97.05{\small$\pm$0.4} & 93.60{\small$\pm$0.1}\\ \bottomrule
\end{tabular}
}
\caption{Average and standard deviation of F1 scores for the best model on the validation sets.}
\label{tab:allmodels}
\end{table}

\begin{table}[ht]
    \setlength{\tabcolsep}{2pt}
    \resizebox{\columnwidth}{!}{%
      \begin{tabular}{@{}lc|ccc|ccc|cccccc|ccc@{}}
        \toprule
        \multicolumn{2}{c|}{\multirow{2}{*}{Evalution set}}          & \multicolumn{3}{c|}{\multirow{2}{*}{French}} & \multicolumn{3}{c|}{\multirow{2}{*}{English}} & \multicolumn{6}{c|}{Multilingual}                                       &  \multicolumn{3}{c}{\multirow{2}{*}{Human Expert}}  \\
        \multicolumn{2}{c|}{}                                        & \multicolumn{3}{c|}{}                        & \multicolumn{3}{c|}{}                         & \multicolumn{3}{c|}{French-Test}  & \multicolumn{3}{c|}{English-Test}   & \multicolumn{3}{c}{}\\
                                                &         & Precision & Recall & F1-Score & Precision & Recall & F1-Score & Precision & Recall & \multicolumn{1}{c|}{F1-Score} & Precision & Recall & \multicolumn{1}{c|}{F1-Score} & Precision & Recall & F1-Score \\ \midrule
        \multirow{2}{*}{\textbf{Full} subset}   & ChatGPT & 0.95      & 1      & 0.97     & 0.99      & 1      & 0.99     & 0.99      & 1      & \multicolumn{1}{c|}{0.99}     & 0.98      & 1      & \multicolumn{1}{c|}{0.99}     & 0.98      & 0.87   & 0.92     \\
                                                & Human   & 1         & 0.94   & 0.97     & 1         & 0.99   & 0.99     & 1         & 0.99   & \multicolumn{1}{c|}{0.99}     & 1         & 0.98   & \multicolumn{1}{c|}{0.99}     & 0.88      & 0.98   & 0.93     \\ \midrule
        \multirow{2}{*}{\textit{+misspelling}} & ChatGPT & 1         & 0.95   & 0.98     & 0.99      & 0.79   & 0.88     & 1         & 0.96   & \multicolumn{1}{c|}{0.98}     & 0.99      & 0.99   & \multicolumn{1}{c|}{0.99}     & -         & -      & -        \\
                                                & Human   & 0.95      & 1      & 0.98     & 0.82      & 0.99   & 0.9      & 0.96      & 1      & \multicolumn{1}{c|}{0.98}     & 0.99      & 0.99   & \multicolumn{1}{c|}{0.99}     & -         & -      & -        \\ \midrule
        \multirow{2}{*}{\textit{+homoglyphs}}  & ChatGPT & 1         & 0.94   & 0.97     & 0.99      & 0.87   & 0.93     & 1         & 0.97   & \multicolumn{1}{c|}{0.99}     & 0.99      & 0.99   & \multicolumn{1}{c|}{0.99}     & -         & -      & -        \\
                                                & Human   & 0.94      & 1      & 0.97     & 0.88      & 0.99   & 0.93     & 0.99      & 0.99   & \multicolumn{1}{c|}{0.99}     & 0.97      & 1      & \multicolumn{1}{c|}{0.99}     & -         & -      & -        \\ \bottomrule
      \end{tabular}
    }
    \caption{Detailed test set scores (full subset) breakdown of \camemberta (French), RoBERTa (English), XLM-R (Multilingual) trained on the full subset. }
    \label{tab:detailedf1score}
\end{table}

\subsubsection{Out-of-Domain}

To assess the potential for overfitting to our in-domain data, we evaluated the performance of our French detector on the out-of-domain test sets described previously. 
The results, shown in Table~\ref{tab:outofdomain}, reveal the detector's exceptional performance on the FTB and FAQ-Gouv test sets, with a drop in accuracy to 88.75 on the FAQ-Rand subset. 
This suggests the model may be detecting translation artifacts that remain in some FAQ web pages after automatic translation. 
Remarkably, our detector correctly identified French text generated natively by ChatGPT, suggesting that it may be possible to develop detectors for other languages by translating existing datasets. 
Similarly, the detector models displayed surprising performance in detecting content generated by BingGPT.

The multilingual detector model consistently outperformed the monolingual model only in detecting human-generated text but fell behind in detecting ChatGPT or BingGPT-generated text, this behavior might be due to the significantly larger pre-training dataset of XLM-R compared to \camemberta.

The detector models also exhibited clear weaknesses against misspelling and homoglyph-based attacks. 
For instance, the performance of \camemberta and XLM-R dropped to 44.81 and 28.18, respectively, when detecting BingGPT-generated text with misspellings added.

Finally, the low scores obtained by the detector on the adversarial response dataset we developed in the style of ChatGPT and BingGPT serve to validate our detector's heavy reliance on the writing style utilized in generating responses.

\begin{table}[ht]
\setlength{\tabcolsep}{3pt}
\resizebox{\columnwidth}{!}{%
\begin{tabular}{@{}c|cccccccccccc|cccccc@{}}
\toprule
True label & \multicolumn{12}{c|}{Human} & \multicolumn{6}{c}{ChatGPT} \\ \midrule
\multirow{2}{*}{Model} & \multicolumn{3}{c|}{FTB} & \multicolumn{3}{c|}{FAQ-Rand} & \multicolumn{3}{c|}{FAQ-Gouv} & \multicolumn{3}{c|}{Adversarial} & \multicolumn{3}{c|}{Native} & \multicolumn{3}{c}{BingGPT} \\
 & raw & \textit{+ms} & \multicolumn{1}{c|}{\textit{+hg}} & raw & \textit{+ms} & \multicolumn{1}{c|}{\textit{+hg}} & \textit{raw} & \textit{+ms} & \multicolumn{1}{c|}{\textit{+hg}} & \textit{raw} & \textit{+ms} & \textit{+hg} & raw & \textit{+ms} & \multicolumn{1}{c|}{\textit{+hg}} & raw & \textit{+ms} & \textit{+hg} \\ \midrule
\camemberta & 99.19 & \textbf{99.92} & \multicolumn{1}{c|}{\textbf{100}} & 88.75 & 99.01 & \multicolumn{1}{c|}{99.10} & 96.17 & \textbf{100} & \multicolumn{1}{c|}{\textbf{99.57}} & 33.57 & 87.61 & \textbf{85.49} & 99.19 & 81.42 & \multicolumn{1}{c|}{84.96} & \textbf{92.45} & 44.81 & 48.37 \\ \midrule
XLM-R & \textbf{99.43} & 99.59 & \multicolumn{1}{c|}{99.76} &  \textbf{95.35} & \textbf{99.39} & \multicolumn{1}{c|}{\textbf{99.55}} & \textbf{96.59} & \textbf{100} & \multicolumn{1}{c|}{\textbf{99.57}} & 59.12 & \textbf{89.05} & 82.67 & 94.69 & 60.18 & \multicolumn{1}{c|}{62.83} & 77.46 & 28.18 & 35.72 \\ \midrule
 \multicolumn{19}{c}{\textit{Trained on a mix of raw, misspellings and homoglyphs$^\ast$}} \\ \midrule
\camemberta & 98.98 & 98.54 & \multicolumn{1}{c|}{98.79} & 80.56 & 84.51 & \multicolumn{1}{c|}{84.73} &  90.64 & 91.49 & \multicolumn{1}{c|}{90.21} & 45.90 & 42.62 & 44.26 & {\bf 100} & {\bf 99.12} & \multicolumn{1}{c|}{{\bf 99.95}} & 91.51 & {\bf 91.51} & {\bf 90.57} \\ \midrule
XLM-R & 98.54 & 98.78 & \multicolumn{1}{c|}{98.79} &  85.20 & 88.84 & \multicolumn{1}{c|}{95.32} & 92.34 & 96.17 & \multicolumn{1}{c|}{ 95.32} & \textbf{62.26} & 60.66 & 62.30 & 100 & 97.34 & \multicolumn{1}{c|}{99.16} & 62.26 & 53.77 & 56.60 \\ \bottomrule
\end{tabular}%
}
\caption{Accuracy scores of \camemberta and XLM-R on the French out-of-domain test sets. \textit{ms}: misspelling, \textit{hg}: homoglyphs. \textit{$^\ast$Dataset mix was 100\% raw, 50\% misspellings and 50\% homoglyphs.}}
\label{tab:outofdomain}
\end{table}

\section{Discussion}
\paragraph{About the link between translation quality and the model detectability}  As part of our study to assess the possibility of differentiating between texts written by humans and those generated by LLMs, following the work of~\citet{guo2023close}, we analyzed and re-evaluated the responses in the translated French dataset.
The purpose was to confirm the hypothesis that a human expert can generally distinguish between a ChatGPT-generated text and one written by a human.
We initially rated the translation quality on a scale of 1 to 5, with 5 indicating a good translation. Translations with scores exceeding 3 were retained even though ChatGPT managed to interpret badly translated questions extremely well.

Additionally, we assessed the correlation between our detector's performance and translation quality scores, and found it to be weak.\footnote{ With three different correlation measures showing the same trend: Spearman's $\tau$ of -0.25, Pearson's $R$ of -0.26 and Kendall's $\tau$  of -0.24.}

\paragraph{About discriminating linguistics clues} We identified several visible characteristics in the generated texts. 
ChatGPT uses an impersonal and didactic style, characterized by extensive use of the impersonal form, conditionals statements, as shown here:
\begin{quote}
\begin{itemize}
    \item ``Cela \textbf{pourrait} également nuire à la réputation de l'entreprise (...)''
    \item ``Cela \textbf{pourrait} entraîner une baisse des dépenses des consommateurs (...)''
    \item ``\textbf{Si} vous êtes allergique aux chats, cela signifie que votre corps a une réaction anormale aux protéines présentes dans leur peau, leur urine ou leur salive. \textbf{Si} vous deviez manger de la viande de chat, il est possible que (...)''
\end{itemize}
\end{quote}

The language model structures its responses to create an impression of coherence and clarity. 
It often reformulates the question in its answer, resulting in a didactic response that aligns with the question. 
\begin{quote}

\begin{itemize}
\item Question: ``Pourquoi {\bf mon signal wifi semble se dégrader avec le temps}~? Je réinitialise/redémarre constamment mon routeur et/ou mon modem. Je dois noter que je vis dans un petit appartement et que j'ai utilisé 2 routeurs haut de gamme. Explique comme si j'avais cinq ans.''\\
GPT : ``Il peut y avoir plusieurs raisons pour lesquelles {\bf votre signal Wi-Fi se dégrade avec le temps}. Voici quelques explications possibles~: (...)''
\end{itemize}
\end{quote}

ChatGPT's responses are often general, and it redefines the subject on which the question is asked.
\begin{quote}
When asked ``How does nature solve for Pi~?'' or ``*Comment la nature résout-elle pour Pi~?'', it started by stating the definition of Pi:
\begin{itemize}
    \item ``Pi, ou le nombre 3,14, est une constante mathématique qui représente le rapport de la circonférence d'un cercle à son diamètre. La valeur de Pi est d'environ 3,14, mais c'est un nombre irrationnel (...)''
\end{itemize}
\end{quote}

Additionally, ChatGPT is characterized by the absence of some human markers, such as errors in punctuation, spelling, or grammar. 
The language model does not use any tone, judgment, or personal touch, such as (``je pense que'' / ``je juge que'') , which creates a neutral impression. 
While its responses lack a human touch, it provides a specific recommendation when discussing technical or sensitive issues, such as consulting a specialist or seeking medical attention.
Also, It does not ask any questions except towards the end of the response.
\begin{quote}
\begin{itemize}
    \item ``(...) \textbf{il est important de consulter un médecin} dès que possible. \textbf{N'hésitez pas à appeler le} 911 ou votre numéro d'urgence local si vous ressentez des douleurs à la poitrine ou d'autres symptômes d'une condition médicale grave.''
    \item ``(...) \textbf{Il est important de} vérifier les instructions de votre four à micro-ondes pour voir si le support en métal peut être utilisé en toute sécurité.''
    \item ``(...) Encore une fois, \textbf{je vous recommande de} parler avec un dermatologue ou un autre professionnel de la santé pour déterminer le plan de traitement le plus approprié pour votre cas spécifique.''
    \item ``(...) J'espère que cela vous aidera à l'expliquer ! \textbf{Y a-t-il autre chose que vous aimeriez savoir sur Vénus ou sur la façon dont elle se déplace dans l'espace~?}''

\end{itemize}
\end{quote}
Our study suggests that these visible differences could be used to differentiate between human-written texts and AI-generated texts automatically. It shall be  noted that the ChatGPT tendency to produce didactic text can lead any detector trained on its content to be easily fooled assuming  the text follows the same patterns. This is what showed our results in Table~\ref{tab:outofdomain} (``Adversarial'' column results).

\paragraph{About the character-level perturbations}  Interestingly, the introduction of character-level perturbations increased the model's capability of detecting the adversarial human content, albeit at the expense of its capacity to detect Bing automatically generated content.
This finding suggests that the addition of perturbations to content renders it more comparable to human-generated content, confirming, for French, previous work on the subject~\citep{wolff2020attacking}.
These effects were much more difficult to notice in the in-domain scenarios because of the high-accuracy of the model.

\paragraph{Enhancing the robustness to noise of our models} Although not the focus of this work, one obvious path of improvement is to add the same kind of perturbations to the training data in order to make the model more robust. To this end, we performed a quick set of experiments where we added to the training set, 50\% of its content perturbed by misspellings and 50\% with homoglyphs leading to a training set twice as big.\footnote{We also tested a 50\% original training set~+~25\% misspelling~+~25\% homoglyphs perturbations model that led to slightly inferior performance, less than one percentage point of difference.}
{ These results, presented in the lower half of Table~\ref{tab:outofdomain}, demonstrate that both models exhibit a minor decrease in human detection accuracy. However, they achieve substantial enhancements and improved robustness, particularly when utilizing \camemberta, for detecting ChatGPT-generated text in the presence of noisy data. Consequently, the detector models are now less inclined to attribute writing errors to human authors and instead focus more on writing style. This is evident from the scores obtained on the Adversarial set, where the performance on noisy data aligns more closely with that on the original set.}
However, this does not make the model less sensitive to other kinds of noises but it is an interesting path of improvement. As always with noisy adversarial user-generated content, the question is to find a more general approach that will avoid a constant {\em cat and mouse} game when it comes to processing productive content.

\paragraph{Take home message} The key takeaway from our study is that detecting adversarial text, which is designed to evade detection by language models, presents a significant challenge. 
OpenAI has reported\footnote{\href{https://openai.com/blog/new-ai-classifier-for-indicating-ai-written-text}{https://openai.com/blog/new-ai-classifier-for-indicating-ai-written-text}} a success rate of 26\% in their own supervised settings when identifying adversarial content in a challenge set of English text.\footnote{Not released at the time of writing.}  Furthermore, OpenAI has stated that their detection methods are unreliable for text shorter than 1000 characters. This is further confirmed by the recent, at the time of writing, findings of \citet{sadasivan2023can} who provided  empirical and theoretical evidence demonstrating the unreliability of existing detection methods in practical scenarios, reinforcing the significant challenge of detecting adversarial text. Notably, \citet{sadasivan2023can} introduced a theoretical impossibility result, which suggests that even the best-possible detector can only achieve marginal performance improvement over a random classifier when facing a sufficiently good language model. 
Their finding confirms our results on the inherent difficulty in reliably detecting AI-generated text.

Thus, we would like to emphasize that our study does not claim to have produced a universally accurate detector. Our strong results are based on in-domain testing and, unsurprisingly, do not generalize in out-of-domain scenarios. This is even more so when used on text specifically designed to fool language model detectors and on text intentionally stylistically similar to ChatGPT-generated text, especially instructional text.\\

\section{Conclusion}
In conclusion, this paper proposed a methodology for developing and evaluating ChatGPT detectors in multiple languages, focusing on French as a case study. 
The proposed method involved translating an English dataset into French and training a classifier on the translated data. 
The results demonstrate that the proposed method can effectively detect ChatGPT-generated text, with a certain degree of robustness against basic attack techniques, albeit exclusively within the in-domain setting. 
However, the detectors display evident vulnerabilities in out-of-domain contexts, emphasizing the importance of considering different writing styles in training language models.
Additionally, the study highlights the significant challenge of detecting adversarial text, which even OpenAI's detection methods have difficulties with. 
The key takeaway is that caution should be exercised when applying in-domain testing results to a wider variety of content. 
We provide \href{https://gitlab.inria.fr/wantoun/robust-chatgpt-detection}{open-source resources} to further advance research in this and are currently working to extend the adversarial dataset to better understand the limitations of these models.

\section*{Acknowledgments}
We thank the reviewers for their insightful comments. This work was partly funded by Benoît Sagot's chair in the PRAIRIE institute funded by the French national research agency (ANR as part of the ``Investissements d’avenir'' program under the reference \mbox{ANR-19-P3IA-0001}). 
This work also received funding from the European Union’s Horizon 2020 research and innovation program under grant agreement No. 101021607. 
The authors are grateful to the OPAL infrastructure from Université Côte d'Azur for providing resources and support.

\bibliographystyle{taln2023}
\bibliography{anthology,custom}

\begin{thebibliography}{~~~}

\bibitem[\protect\citename{Abeill{\'e} {\em et~al.},
  }2000]{abeille-etal-2000-building}
{\sc Abeill{\'e} A., Cl{\'e}ment L. \& Kinyon A.} (2000).
\newblock Building a treebank for {F}rench.
\newblock In {\em Proceedings of the Second International Conference on
  Language Resources and Evaluation ({LREC}{'}00)}, Athens, Greece: European
  Language Resources Association (ELRA).

\bibitem[\protect\citename{Antoun {\em et~al.},
  }2021a]{antoun-etal-2021-araelectra}
{\sc Antoun W., Baly F. \& Hajj H.} (2021a).
\newblock {A}ra{ELECTRA}: Pre-training text discriminators for {A}rabic
  language understanding.
\newblock In {\em Proceedings of the Sixth Arabic Natural Language Processing
  Workshop}, p.\ 191--195, Kyiv, Ukraine (Virtual): Association for
  Computational Linguistics.

\bibitem[\protect\citename{Antoun {\em et~al.},
  }2021b]{antoun-etal-2021-aragpt2}
{\sc Antoun W., Baly F. \& Hajj H.} (2021b).
\newblock {A}ra{GPT}2: Pre-trained transformer for {A}rabic language
  generation.
\newblock In {\em Proceedings of the Sixth Arabic Natural Language Processing
  Workshop}, p.\ 196--207, Kyiv, Ukraine (Virtual): Association for
  Computational Linguistics.

\bibitem[\protect\citename{Antoun {\em et~al.}, }2023]{antoun2023data}
{\sc Antoun W., Sagot B. \& Seddah D.} (2023).
\newblock Data-efficient french language modeling with camemberta.
\newblock In {\em Findings of the Association for Computational Linguistics:
  ACL 2023}, Toronto, Canada: Association for Computational Linguistics.

\bibitem[\protect\citename{Bender {\em et~al.}, }2021]{bender2021onthedangers}
{\sc Bender E.~M., Gebru T., McMillan-Major A. \& Shmitchell S.} (2021).
\newblock On the dangers of stochastic parrots: Can language models be too big?
\newblock \doi{10.1145/3442188.3445922}.

\bibitem[\protect\citename{Brown {\em et~al.}, }2020]{brown2020language}
{\sc Brown T., Mann B., Ryder N., Subbiah M., Kaplan J.~D., Dhariwal P.,
  Neelakantan A., Shyam P., Sastry G., Askell A. {\em et~al.}} (2020).
\newblock Language models are few-shot learners.
\newblock {\em Advances in neural information processing systems}, {\bf 33},
  1877--1901.

\bibitem[\protect\citename{Chen {\em et~al.}, }2020]{chen2020MedDialog-en-zh}
{\sc Chen S., Ju Z., Dong X., Fang H., Wang S., Yang Y., Zeng J., Zhang R.,
  Zhang R., Zhou M., Zhu P. \& Xie P.} (2020).
\newblock Meddialog: a large-scale medical dialogue dataset.
\newblock {\em arXiv preprint arXiv:2004.03329}.

\bibitem[\protect\citename{Chowdhery {\em et~al.}, }2022]{chowdhery2022palm}
{\sc Chowdhery A., Narang S., Devlin J., Bosma M., Mishra G., Roberts A.,
  Barham P., Chung H.~W., Sutton C., Gehrmann S. {\em et~al.}} (2022).
\newblock Palm: Scaling language modeling with pathways.
\newblock {\em arXiv preprint arXiv:2204.02311}.

\bibitem[\protect\citename{Christiano {\em et~al.}, }2017]{christiano2017deep}
{\sc Christiano P.~F., Leike J., Brown T., Martic M., Legg S. \& Amodei D.}
  (2017).
\newblock Deep reinforcement learning from human preferences.
\newblock In {\sc I. Guyon, U.~V. Luxburg, S. Bengio, H. Wallach, R. Fergus, S.
  Vishwanathan \& R. Garnett}, \'Eds., {\em Advances in Neural Information
  Processing Systems}, volume~30: Curran Associates, Inc.

\bibitem[\protect\citename{Clark {\em et~al.}, }2020]{clark2020electra}
{\sc Clark K., Luong M.-T., Le Q.~V. \& Manning C.~D.} (2020).
\newblock {ELECTRA}: Pre-training text encoders as discriminators rather than
  generators.
\newblock In {\em ICLR}.

\bibitem[\protect\citename{Conneau {\em et~al.},
  }2020]{conneau-etal-2020-unsupervised}
{\sc Conneau A., Khandelwal K., Goyal N., Chaudhary V., Wenzek G., Guzm{\'a}n
  F., Grave E., Ott M., Zettlemoyer L. \& Stoyanov V.} (2020).
\newblock Unsupervised cross-lingual representation learning at scale.
\newblock In {\em Proceedings of the 58th Annual Meeting of the Association for
  Computational Linguistics}, p.\ 8440--8451, Online: Association for
  Computational Linguistics.
\newblock \doi{10.18653/v1/2020.acl-main.747}.

\bibitem[\protect\citename{De~Bruyn {\em et~al.},
  }2021]{de-bruyn-etal-2021-mfaq}
{\sc De~Bruyn M., Lotfi E., Buhmann J. \& Daelemans W.} (2021).
\newblock {MFAQ}: a multilingual {FAQ} dataset.
\newblock In {\em Proceedings of the 3rd Workshop on Machine Reading for
  Question Answering}, p.\ 1--13, Punta Cana, Dominican Republic: Association
  for Computational Linguistics.
\newblock \doi{10.18653/v1/2021.mrqa-1.1}.

\bibitem[\protect\citename{Devlin {\em et~al.}, }2019]{devlin-etal-2019-bert}
{\sc Devlin J., Chang M.-W., Lee K. \& Toutanova K.} (2019).
\newblock {BERT}: Pre-training of deep bidirectional transformers for language
  understanding.
\newblock In {\em Proceedings of the 2019 Conference of the North {A}merican
  Chapter of the Association for Computational Linguistics: Human Language
  Technologies, Volume 1 (Long and Short Papers)}, p.\ 4171--4186, Minneapolis,
  Minnesota: Association for Computational Linguistics.
\newblock \doi{10.18653/v1/N19-1423}.

\bibitem[\protect\citename{Fagni {\em et~al.}, }2021]{fagni2021tweepfake}
{\sc Fagni T., Falchi F., Gambini M., Martella A. \& Tesconi M.} (2021).
\newblock Tweepfake: About detecting deepfake tweets.
\newblock {\em Plos one}, {\bf 16}(5), e0251415.

\bibitem[\protect\citename{Fan {\em et~al.}, }2019]{reddit-eli5_lfqa}
{\sc Fan A., Jernite Y., Perez E., Grangier D., Weston J. \& Auli M.} (2019).
\newblock {ELI5:} long form question answering.
\newblock In {\sc A. Korhonen, D.~R. Traum \& L. M{\`{a}}rquez}, \'Eds., {\em
  Proceedings of the 57th Conference of the Association for Computational
  Linguistics, {ACL} 2019, Florence, Italy, July 28- August 2, 2019, Volume 1:
  Long Papers}, p.\ 3558--3567: Association for Computational Linguistics.
\newblock \doi{10.18653/v1/p19-1346}.

\bibitem[\protect\citename{Fedus {\em et~al.}, }2021]{fedus2021switch}
{\sc Fedus W., Zoph B. \& Shazeer N.} (2021).
\newblock Switch transformers: Scaling to trillion parameter models with simple
  and efficient sparsity.
\newblock {\em arXiv preprint arXiv:2101.03961}.

\bibitem[\protect\citename{Guo {\em et~al.}, }2023]{guo2023close}
{\sc Guo B., Zhang X., Wang Z., Jiang M., Nie J., Ding Y., Yue J. \& Wu Y.}
  (2023).
\newblock How close is chatgpt to human experts? comparison corpus, evaluation,
  and detection.
\newblock {\em arXiv preprint arXiv:2301.07597}.

\bibitem[\protect\citename{He {\em et~al.}, }2021]{he2021debertav3}
{\sc He P., Gao J. \& Chen W.} (2021).
\newblock Debertav3: Improving deberta using electra-style pre-training with
  gradient-disentangled embedding sharing.

\bibitem[\protect\citename{Hoffmann {\em et~al.}, }2022]{hoffmann2022training}
{\sc Hoffmann J., Borgeaud S., Mensch A., Buchatskaya E., Cai T., Rutherford
  E., Casas D. d.~L., Hendricks L.~A., Welbl J., Clark A. {\em et~al.}} (2022).
\newblock Training compute-optimal large language models.
\newblock {\em arXiv preprint arXiv:2203.15556}.

\bibitem[\protect\citename{Jawahar {\em et~al.},
  }2022]{jawahar-etal-2022-automatic}
{\sc Jawahar G., Abdul-Mageed M. \& Lakshmanan L.} (2022).
\newblock Automatic detection of entity-manipulated text using factual
  knowledge.
\newblock In {\em Proceedings of the 60th Annual Meeting of the Association for
  Computational Linguistics (Volume 2: Short Papers)}, p.\ 86--93, Dublin,
  Ireland: Association for Computational Linguistics.
\newblock \doi{10.18653/v1/2022.acl-short.10}.

\bibitem[\protect\citename{Liu {\em et~al.}, }2019]{liu2019roberta}
{\sc Liu Y., Ott M., Goyal N., Du J., Joshi M., Chen D., Levy O., Lewis M.,
  Zettlemoyer L. \& Stoyanov V.} (2019).
\newblock Roberta: A robustly optimized bert pretraining approach.
\newblock {\em arXiv preprint arXiv:1907.11692}.

\bibitem[\protect\citename{Ma, }2019]{ma2019nlpaug}
{\sc Ma E.} (2019).
\newblock Nlp augmentation.
\newblock https://github.com/makcedward/nlpaug.

\bibitem[\protect\citename{Maia {\em et~al.}, }2018]{fiqa-2018}
{\sc Maia M., Handschuh S., Freitas A., Davis B., McDermott R., Zarrouk M. \&
  Balahur A.} (2018).
\newblock Www'18 open challenge: Financial opinion mining and question
  answering.
\newblock In {\em Companion Proceedings of The Web Conference 2018}, WWW '18,
  p.\ 1941–1942, Republic and Canton of Geneva, CHE: International World Wide
  Web Conferences Steering Committee.
\newblock \doi{10.1145/3184558.3192301}.

\bibitem[\protect\citename{Martin {\em et~al.},
  }2020]{martin-etal-2020-camembert}
{\sc Martin L., Muller B., Ortiz~Su{\'a}rez P.~J., Dupont Y., Romary L., de~la
  Clergerie {\'E}., Seddah D. \& Sagot B.} (2020).
\newblock {C}amem{BERT}: a tasty {F}rench language model.
\newblock In {\em Proceedings of the 58th Annual Meeting of the Association for
  Computational Linguistics}, p.\ 7203--7219, Online: Association for
  Computational Linguistics.
\newblock \doi{10.18653/v1/2020.acl-main.645}.

\bibitem[\protect\citename{McGuffie \& Newhouse,
  }2020]{mcguffie2020radicalization}
{\sc McGuffie K. \& Newhouse A.} (2020).
\newblock The radicalization risks of gpt-3 and advanced neural language
  models.
\newblock {\em arXiv preprint arXiv:2009.06807}.

\bibitem[\protect\citename{Mitchell {\em et~al.}, }2023]{mitchell2023detectgpt}
{\sc Mitchell E., Lee Y., Khazatsky A., Manning C.~D. \& Finn C.} (2023).
\newblock Detectgpt: Zero-shot machine-generated text detection using
  probability curvature.
\newblock {\em arXiv preprint arXiv:2301.11305}.

\bibitem[\protect\citename{Nguyen-Son {\em et~al.}, }2021]{nguyen2021machine}
{\sc Nguyen-Son H.-Q., Thao T., Hidano S., Gupta I. \& Kiyomoto S.} (2021).
\newblock Machine translated text detection through text similarity with
  round-trip translation.
\newblock In {\em Proceedings of the 2021 Conference of the North American
  Chapter of the Association for Computational Linguistics: Human Language
  Technologies}, p.\ 5792--5797.

\bibitem[\protect\citename{Ouyang {\em et~al.}, }2022]{ouyang2022training}
{\sc Ouyang L., Wu J., Jiang X., Almeida D., Wainwright C., Mishkin P., Zhang
  C., Agarwal S., Slama K., Gray A., Schulman J., Hilton J., Kelton F., Miller
  L., Simens M., Askell A., Welinder P., Christiano P., Leike J. \& Lowe R.}
  (2022).
\newblock Training language models to follow instructions with human feedback.
\newblock In {\sc A.~H. Oh, A. Agarwal, D. Belgrave \& K. Cho}, \'Eds., {\em
  Advances in Neural Information Processing Systems}.

\bibitem[\protect\citename{Radford {\em et~al.}, }2018]{radford2018improving}
{\sc Radford A., Narasimhan K., Salimans T. \& Sutskever I.} (2018).
\newblock Improving language understanding by generative pre-training.

\bibitem[\protect\citename{Radford {\em et~al.}, }2019]{radford2019language}
{\sc Radford A., Wu J., Child R., Luan D., Amodei D. \& Sutskever I.} (2019).
\newblock Language models are unsupervised multitask learners.

\bibitem[\protect\citename{Rae {\em et~al.}, }2021]{rae2021scaling}
{\sc Rae J.~W., Borgeaud S., Cai T., Millican K., Hoffmann J., Song F.,
  Aslanides J., Henderson S., Ring R., Young S. {\em et~al.}} (2021).
\newblock Scaling language models: Methods, analysis \& insights from training
  gopher.
\newblock {\em arXiv preprint arXiv:2112.11446}.

\bibitem[\protect\citename{Raffel {\em et~al.}, }2020]{raffel2019exploring}
{\sc Raffel C., Shazeer N., Roberts A., Lee K., Narang S., Matena M., Zhou Y.,
  Li W. \& Liu P.~J.} (2020).
\newblock Exploring the limits of transfer learning with a unified text-to-text
  transformer.
\newblock {\em Journal of Machine Learning Research}, {\bf 21}(140), 1--67.

\bibitem[\protect\citename{Sadasivan {\em et~al.}, }2023]{sadasivan2023can}
{\sc Sadasivan V.~S., Kumar A., Balasubramanian S., Wang W. \& Feizi S.}
  (2023).
\newblock Can ai-generated text be reliably detected?
\newblock {\em arXiv preprint arXiv:2303.11156}.

\bibitem[\protect\citename{Shoeybi {\em et~al.}, }2019]{shoeybi2019megatron}
{\sc Shoeybi M., Patwary M., Puri R., LeGresley P., Casper J. \& Catanzaro B.}
  (2019).
\newblock Megatron-{LM}: Training multi-billion parameter language models using
  model parallelism.
\newblock {\em arXiv preprint arXiv:1909.08053}.

\bibitem[\protect\citename{Solaiman {\em et~al.}, }2019]{solaiman2019release}
{\sc Solaiman I., Brundage M., Clark J., Askell A., Herbert-Voss A., Wu J.,
  Radford A., Krueger G., Kim J.~W., Kreps S. {\em et~al.}} (2019).
\newblock Release strategies and the social impacts of language models.
\newblock {\em arXiv preprint arXiv:1908.09203}.

\bibitem[\protect\citename{Stiennon {\em et~al.}, }2020]{Stiennon2020learning}
{\sc Stiennon N., Ouyang L., Wu J., Ziegler D., Lowe R., Voss C., Radford A.,
  Amodei D. \& Christiano P.~F.} (2020).
\newblock Learning to summarize with human feedback.
\newblock In {\sc H. Larochelle, M. Ranzato, R. Hadsell, M. Balcan \& H. Lin},
  \'Eds., {\em Advances in Neural Information Processing Systems}, volume~33,
  p.\ 3008--3021: Curran Associates, Inc.

\bibitem[\protect\citename{Uchendu {\em et~al.},
  }2020]{uchendu-etal-2020-authorship}
{\sc Uchendu A., Le T., Shu K. \& Lee D.} (2020).
\newblock Authorship attribution for neural text generation.
\newblock In {\em Proceedings of the 2020 Conference on Empirical Methods in
  Natural Language Processing (EMNLP)}, p.\ 8384--8395, Online: Association for
  Computational Linguistics.
\newblock \doi{10.18653/v1/2020.emnlp-main.673}.

\bibitem[\protect\citename{Weidinger {\em et~al.}, }2021]{weidinger2021ethical}
{\sc Weidinger L., Mellor J., Rauh M., Griffin C., Uesato J., Huang P.-S.,
  Cheng M., Glaese M., Balle B., Kasirzadeh A. {\em et~al.}} (2021).
\newblock Ethical and social risks of harm from language models.
\newblock {\em arXiv preprint arXiv:2112.04359}.

\bibitem[\protect\citename{Wolff \& Wolff, }2020]{wolff2020attacking}
{\sc Wolff M. \& Wolff S.} (2020).
\newblock Attacking neural text detectors.
\newblock {\em arXiv preprint arXiv:2002.11768}.

\bibitem[\protect\citename{Yang {\em et~al.}, }2015]{yang2015wikiqa}
{\sc Yang Y., Yih S. W.-t. \& Meek C.} (2015).
\newblock Wikiqa: A challenge dataset for open-domain question answering.
\newblock In {\em Proceedings of the 2015 Conference on Empirical Methods in
  Natural Language Processing}: ACL - Association for Computational
  Linguistics.

\bibitem[\protect\citename{Zellers {\em et~al.}, }2019]{zellers2019defending}
{\sc Zellers R., Holtzman A., Rashkin H., Bisk Y., Farhadi A., Roesner F. \&
  Choi Y.} (2019).
\newblock Defending against neural fake news.
\newblock {\em Advances in neural information processing systems}, {\bf 32}.

\end{thebibliography}

\end{document}